\documentclass[runningheads]{llncs}
\usepackage{graphicx}
\usepackage{comment}
\usepackage{amsmath,amssymb} 
\usepackage{color}
\usepackage{multirow}

\usepackage{xspace}
\makeatletter
\DeclareRobustCommand\onedot{\futurelet\@let@token\@onedot}
\def\@onedot{\ifx\@let@token.\else.\null\fi\xspace}

\def\eg{\emph{e.g}\onedot}

\def\etc{\emph{etc}\onedot}

\makeatother

\begin{document}
\pagestyle{headings}
\mainmatter
\def\ECCVSubNumber{****}  

\title{MobilePose: Real-Time Pose Estimation for Unseen Objects with Weak Shape Supervision} 

\titlerunning{MobilePose}
%
\author{Tingbo Hou \and
Adel Ahmadyan \and
Liangkai Zhang \and
Jianing Wei \and
Matthias Grundmann}
\authorrunning{T. Hou et al.}
%
\institute{Google Research, 1600 Amphitheatre Pkwy, Mountain View, CA 94043
\email{\{tingbo,ahmadyan,liangkai,jianingwei,grundman\}@google.com}}
\maketitle

\begin{abstract}
In this paper, we address the problem of detecting unseen objects from RGB images and estimating their poses in 3D. We propose two mobile friendly networks: MobilePose-Base and MobilePose-Shape. The former is used when there is only pose supervision, and the latter is for the case when shape supervision is available, even a weak one. We revisit shape features used in previous methods, including segmentation and coordinate map. We explain when and why pixel-level shape supervision can improve pose estimation. Consequently, we add shape prediction as an intermediate layer in the MobilePose-Shape, and let the network learn pose from shape. Our models are trained on mixed real and synthetic data, with weak and noisy shape supervision. They are ultra lightweight that can run in real-time on modern mobile devices (\eg 36 FPS on Galaxy S20). Comparing with previous single-shot solutions, our method has higher accuracy, while using a significantly smaller model ($2\sim3\%$ in model size or number of parameters).

\keywords{Pose estimation, 3D detection, shape, segmentation, mobile}
\end{abstract}

\section{Introduction}

Detecting unseen objects from images and estimating their poses in 3D remain a challenge in computer vision, which have not been fully explored in previous work. Solving this problem enables many applications across computer vision, augmented reality (AR), autonomous driving, and robotics. Furthermore, mobile-friendly solutions add their own layers of complexity to the problem: first, they should run in real-time with limited model size; second, unlike self-driving cars where cameras are fixed~\cite{Mousavian_2017_BBox,Dijk_2019_Depth}, on-device models have to cope with various device rotations.

Towards the challenges, existing methods often simplify the problem by assuming objects are known. Most of the methods from literature are instance-aware~\cite{Kehl_2017_SSD6D,Xiang_2018_PoseCNN,Tekin_2018_SingleShot,Sundermeyer_2018_Implicit,Hu_2019_Segmentation}. Meaning that, they are trained on a set of specific objects, and expected to work on the same instances. Object-specific features including appearance and geometry can be learned to determine poses, and hence, applications are mostly limited to grabbing known objects in robotics. Recent progress in pose estimation has been made by leveraging 2D-3D correspondence at inference time~\cite{Li_2018_DeepIM,Zakharov_2019_DPOD,Peng_2019_PVNet,Li_2019_CDPN,Park_2019_Pix2Pose}. These methods require CAD models of the objects, which are not applicable to unseen ones. Recently, there are a few attempts for removing the requirement of known objects. As an example, \cite{Wang_2019_NOCS} uses depth images to align unseen objects at inference time. While relieving prior knowledge of objects, it relies on depth sensors, which requires extra hardware that is not available on general mobile devices.

For unseen objects, we want the model to learn intra-category features, \eg common shape or geometry of a category. We categorize geometry related representations and name them shape features, which can be mapped to images with pixel-level signals. Shape features have been previously used in pose estimation, \eg segmentation~\cite{Rad_2017_BB8,Hu_2019_Segmentation}, parameterization map~\cite{Zakharov_2019_DPOD}, and coordinate map~\cite{Wang_2019_NOCS,Li_2019_CDPN}. These methods train Convolutional Neural Networks (CNNs) to infer shape features, which are then used out of the networks. They rely on highly accurate predictions of shape features, preferably with high resolutions, to align with objects' CAD models. However, estimating shapes of unseen objects is as hard as, or maybe even harder than estimating poses. Instead of post-processing, we use shape prediction as an intermediate layer, and have the network learn pose from shape. Besides, pixel-level shape labels are expensive to annotate, which brings another challenge to the problem. This motivates us to look for weakly supervised solutions.

Although there are methods claiming real-time~\cite{Tekin_2018_SingleShot,Li_2019_CDPN}, none of them haven been deployed to mobile devices. To run on mobile, the CNN model needs to be ultra lightweight, \eg MobileNet~\cite{Howard_2017_MobileNets} and MobileNetv2~\cite{Sandler_2018_MobileNetv2}, which only have a few million parameters. Another requirement is post-processing, whose runtime also counts. This is often overlooked in previous methods~\cite{Peng_2019_PVNet,Zakharov_2019_DPOD,Li_2019_CDPN,Wang_2019_NOCS}, where expensive algorithms are widely used, \eg RANdom Sample Consensus (RANSAC), Perspective-n-Point (PnP), and Iterative Closest Point (ICP).

In this paper, we address the aforementioned challenges and limitations by proposing two mobile-friendly networks: MobilePose-Base and MobilePose-Shape. 
MobilePose-Base is our baseline network with minimal model size, which detects unseen objects and estimates their poses in a single shot. The detection is anchor-free, following a rising trend in 2D object detection~\cite{Law_2018_CornerNet,Duan_2019_CenterNet,Zhou_2019_CenterNet}. It regresses projected vertices of a 3D bounding box to estimate object's pose. Since pose estimation is in a low-dimensional space, high-resolution features need supervision in order to make a positive contribution. Therefore, we also propose MobilePose-Shape, which predicts shape features in an intermediate layer. Running on mobile devices, the two networks only require cheap operations in post-processing to fully recover object's rotation, translation, and size up to a scale. In this work, we are particularly interested in shoes. Following fashion trends, shoes have changing appearances and shapes with a number of sub-categories, \eg sneakers, boots, flip-flops, high heels, \etc.

To summarize, our contributions in this paper are
\begin{itemize}
    \item We propose two novel MobilePose networks for detecting unseen objects from RGB images and estimating their poses. Unlike previous methods, we do not require any prior knowledge of the objects or additional hardware at inference time.
    \item We revisit shape supervision by exploring when and why it can improve pose estimation. Comparing with previous methods that have shape prediction in parallel to other streams and use it in post-processing, we insert it as an intermediate layer and let the network learn pose from shape.
    \item We train models with weak shape supervision, which transfers shape learning from synthetic data to real data. With the lightweight models, we develop end-to-end applications that can run on mobile devices in real-time. 
\end{itemize}

\section{Related Work}
Given a large literature on pose estimation, we categorize recent work by the prior knowledge the methods require at inference time. 

\textbf{Instance-aware} methods learn poses from a set of known objects and are expected to work on the same instances. SSD-6D~\cite{Kehl_2017_SSD6D} extends the classic architecture in 2D object detection to 6DoF pose estimation. It predicts 2D bounding box, discretized viewpoint, and in-plane rotation in a single shot. PoseCNN~\cite{Xiang_2018_PoseCNN} estimates 3D translation of an object as image locations and depth while regressing over 3D rotation. \cite{Tremblay_2018_Dope} predicts the location of the 3D bounding box's vertices in an image using heatmaps, then recovers the orientation and translation using PnP by knowing the object size. In~\cite{Sundermeyer_2018_Implicit}, an Augmented Autoencoder (AAE) is used to extract orientation from other factors, \eg translation and scale. A codebook is created for each object, which contains AAE embeddings of all orientations. YOLO-6D~\cite{Tekin_2018_SingleShot} predicts image locations of projected box vertices, and recovers 6DoF pose using PnP. BB8~\cite{Rad_2017_BB8} uses coarse segmentation to roughly locate objects, subsequently estimating the corners of a 3D bounding box. The method in~\cite{Hu_2019_Segmentation} parallels segmentation and bounding box estimation as two branches of the network. Pose estimation is improved by letting the network learn the entire shape of an object. \cite{Song_2020_Hybrid} utilizes a hybrid intermediate representation to provide more supervision during training.

\textbf{Model-aware} methods require 3D CAD models of objects in post-processing. This is a much stronger prior, which leverages 2D-3D correspondences and yields higher accuracy. PVNet~\cite{Peng_2019_PVNet} finds 2D-3D correspondence of object features, and formulates pose estimation as a PnP problem. By assuming known 3D models of target objects, iterative mechanism~\cite{Li_2018_DeepIM} has been utilized to refine estimated pose by comparing rendered images with inputs. Another recent method~\cite{Zakharov_2019_DPOD} computes the UV map of the object from a single RGB method and uses PnP + RANSAC to estimate the 6DOF object pose. The UV map is a parameterization of 3D models, which also provides 2D-3D correspondence. \cite{Li_2019_CDPN} estimates rotation and translation separately, where rotation is computed, again, by RANSAC + PnP from coordinate map. Similarly, Pix2Pose~\cite{Park_2019_Pix2Pose} also predicts 3D coordinates of objects from images, and uses RANSAC + PnP to recover poses. For a better prediction, it adopts Generative
Adversarial Network (GAN) in training to discriminate predicted coordinates and rendered coordinates.

\textbf{Depth-aware} methods require depth images in addition to RGB images for pose estimation. In~\cite{Kehl_2016_RGBD}, pose is estimated by searching over the nearest neighbors in a codebook of encoded RGB-D patches. DenseFusion~\cite{Wang_2019_DenseFusion} processes the RGB image and depth image individually, and uses a dense fusion network to extract pixel-wise dense features. Since it has 3D coordinates, it directly predicts translation and rotation. In~\cite{Li_2018_MVMC}, a multi-view framework was proposed using viewpoint alignment and pose voting. It adopts depth image as an optional input. A recent work~\cite{Wang_2019_NOCS} predicts object's normalized coordinates, and aligns them with a depth image to compute the pose. The normalized coordinates compose yet another 2D-3D correspondence mapping.

\textbf{Detection-aware} methods rely on existing 2D detectors to find a bounding box or ROI of an object. \cite{Mousavian_2017_BBox} estimates 3D bounding boxes of vehicles by applying geometric constraints on 2D bounding boxes. The SSD-6D also detects 2D bounding boxes using a SSD-style network. \cite{Wang_2019_NOCS} employees the Mask R-CNN~\cite{He_2017_MaskRCNN} to detect objects and find their 2D locations. \cite{Li_2019_CDPN} adopts YOLOv3 for 2D detector to crop images with objects during inference. Pix2Pose~\cite{Park_2019_Pix2Pose} also assumes cropped images of objects, which are obtained by a modified Fast R-CNN~\cite{Ren_2015_Faster} and Retinanet~\cite{Lin_2017_Focal}.

\textbf{Shape features} have been employed in estimating poses, \eg segmentation~\cite{Rad_2017_BB8,Hu_2019_Segmentation}, 3D features~\cite{Peng_2019_PVNet}, parameterization map~\cite{Zakharov_2019_DPOD}, coordinate map~\cite{Wang_2019_NOCS,Li_2019_CDPN,Park_2019_Pix2Pose}, \etc. These methods use shape prediction in post-processing, instead of in network. That is, they first infer shape features by CNN, and then use PnP or ICP to align them with the 3D models. On the contrary, we adopt shape prediction as an intermediate layer, and let the CNN learn 3D bounding boxes from them. This generalizes our method to unseen objects.

\textbf{Real-time} solutions have been proposed to push the technique closer to applications. The models need to be lightweight in order to run in real-time, preferably in a single shot. \cite{Kehl_2017_SSD6D} uses Inceptionv4 as backbone to build a SSD-style architecture. \cite{Tekin_2018_SingleShot,Hu_2019_Segmentation} both adopt YOLOv2~\cite{Redmon_2016_YOLOv2} as backbone. In a recent work~\cite{Li_2019_CDPN}, YOLOv3 is used to detect objects in the first stage, while pose is estimated in the second stage. With the limitations of model size and runtime, none of them has been deployed to mobile devices and runs in real-time there.

\section{MobilePose-Base}
In this section, we propose MobilePose-Base as our baseline network, which detects unseen objects without anchors and estimate their poses in a single shot.

\begin{figure}[t]
    \centering
    \includegraphics[width=0.85\textwidth]{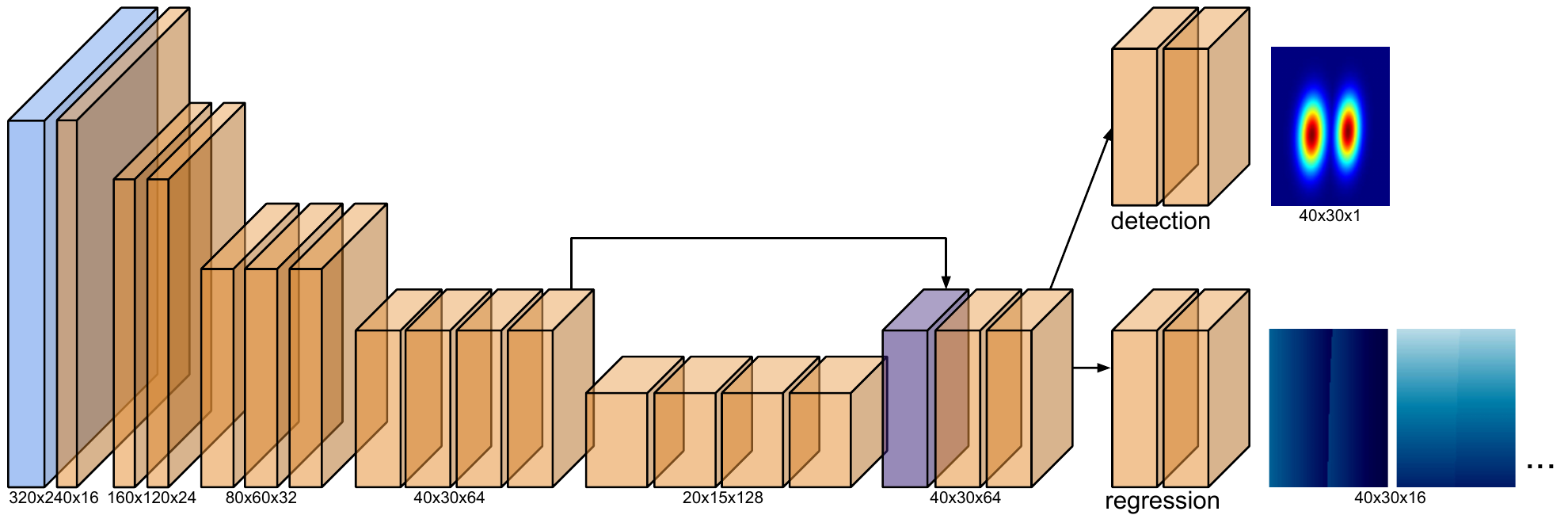}
    \caption{MobilePose-Base network. The blue and purple boxes are convolutional and deconvolutional blocks, respectively. Orange boxes represent inverted residual blocks from MobileNetv2~\cite{Sandler_2018_MobileNetv2}.}
    \label{fig:base}
\end{figure}

\subsection{Backbone}
We devise the backbone as a popular encoder-decoder architecture. To build an ultra lightweight model, we select the MobileNetv2~\cite{Sandler_2018_MobileNetv2} to build our encoder, which is proven to run real-time on mobile, and outperforms YOLOv2~\cite{Redmon_2016_YOLOv2}. The MobileNetv2 is built upon inverted residual blocks, where shortcut connections are between thin bottleneck layers. An expansion-and-squeeze scheme is used in the blocks. To make it even lighter, we remove some blocks with large channels at the bottleneck, reducing half of the parameters.

As shown in Fig.~\ref{fig:base}, the blue and purple boxes are convolutional and deconvolutional blocks, respectively. An orange box represents an inverted residual block. The numbers of blocks and their dimensions shown in the figure are exactly the same in our implementation. The input is an image with size $640\times480\times3$. The encoder starts with a convolutional layer, followed by five levels of inverted residual blocks. At the bottleneck, we use four 128-channel blocks, instead of four 160-channel blocks and one 320-channel block in MobileNetv2~\cite{Sandler_2018_MobileNetv2}. The decoder is composed by a deconvolution layer, a concatenation layer with skip connection from the same scale in the encoder, and two inverted residual blocks.

\subsection{Heads}\label{sec:head}
\begin{figure}[t]
    \centering
    \includegraphics[width=0.2\textwidth]{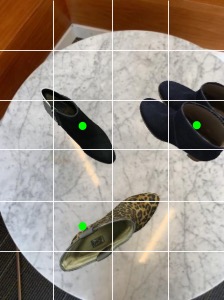}\hspace{5pt}
    \includegraphics[width=0.2\textwidth]{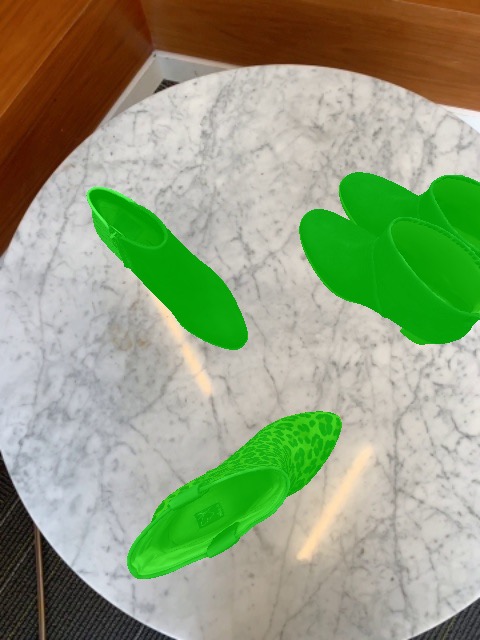}\hspace{5pt}
    \includegraphics[width=0.2\textwidth]{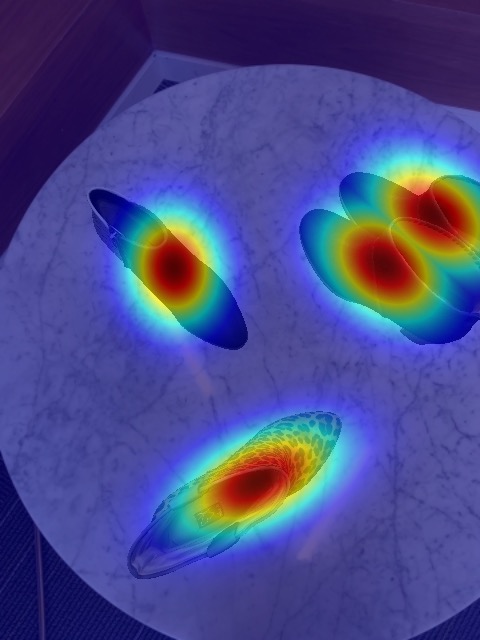}
    \caption{Detection methods (from left to right): anchor, segmentation, and distribution.}
    \label{fig:detection}
\end{figure}

We attach two heads after the backbone: detection and regression. The detection head is inspired by the anchor-free methods~\cite{Law_2018_CornerNet,Zhou_2019_CenterNet} in 2D object detection. We model objects as distributions around their centers. The detection head outputs a $40\times30\times1$ heatmap. Specifically, for image $\mathcal{I}$ with pixels $\{\mathbf{p}\}$, its heatmap is computed as a bivariate normal distribution~\cite{Ding_2019_Distribution} 
\begin{equation}
\label{eq:heat}
\mathcal{H}(\mathbf{p}) = \underset{i \in \mathcal{O}}{\max} (\mathcal{N}(\mathbf{p} - \mu_i, \sigma_i)),
\end{equation}
where $\mathcal{O}$ is the set of all object instances in the image, $\mu_i$ is the centroid location of object $i$, and $\sigma_i$ is the kernel size that is proportional to object size. We keep the fractions when computing projections $\mu_i$ to preserve accuracy. For multiple objects in an image, we select the max heat for each pixel, as the examples shown in Fig.~\ref{fig:base} and Fig.~\ref{fig:detection}. By modeling objects as distributions, we end up using a simple L2 loss (mean squared error) for this head.

In Fig.~\ref{fig:detection}, we compare different detection methods used in single-shot pose estimation. Anchor-based methods (\eg \cite{Tekin_2018_SingleShot}) set anchors at grid cells, and regresses bounding boxes at positive anchors (marked as green dots). It handles multiple objects in the same cell by assigning a number of anchors ad hocly. Segmentation methods (\eg \cite{Hu_2019_Segmentation}) find objects by segmented instances. For multiple objects from the same category, it needs instance segmentation to distinguish objects. We model objects as Gaussian distributions, and detect them by finding peaks. We use a high resolution in the figure for better illustration, while the actual resolution in our model is ($40\times30$).

The regression head estimates displacement fields of bounding box vertices, similar with the vertex offsets in~\cite{Hu_2019_Segmentation}. Specifically, for a box vertex $\mathbf{X}_i$, let $\mathbf{x}_i$ denote its projection on the image plane. We compute the displacement vector as
\begin{equation}
\label{eq:distance}
\mathcal{D}_i(\mathbf{p}) = \mathbf{x}_i - \mathbf{p}. 
\end{equation}
Displacement fields of multiple objects in an image are merged according to their heats, as shown in Fig.~\ref{fig:detection}. The regression head outputs a $40\times30\times16$ tensor, where each box vertex contributes two channels of displacements. In the figure, we only show two displacement fields for illustration. To tolerate errors in peak extraction, we regress displacements at all pixels with significant heats. We use L1 loss (mean absolute error) for this head, which is more robust to outliers. With predicted $\mathcal{D}_i(\mathbf{p})$, the loss is computed as
\begin{equation}
    \mathcal{L}_{reg}= \underset{\mathcal{H}(\mathbf{p})>\epsilon}{\text{mean}}(
    ||\mathcal{D}_i(\mathbf{p}) + \mathbf{p} - \mathbf{x}_i||_1),
\end{equation}
where $||\cdot||_1$ denotes the L1-norm, and $\epsilon$ is a threshold (0.2 in our implementation).

\section{MobilePose-Shape}
When shape supervision is available, even a weak one, we provide MobilePose-Shape, which predicts shape features at an intermediate layer.

\subsection{Shape Features}
The motivation is to guide the network learn high-resolution shape features that are related to pose estimation. We found that simply introducing high-resolution features without supervision does not improve our pose estimation. This is because the regression of bounding box vertices is in a low dimensional space. Without supervision, the network may overfit on object-specific features at small scales. The problem is not valid for instance-aware pose estimation, but not our case of unseen objects. 

\begin{figure}[t]
    \centering
    \includegraphics[width=\textwidth]{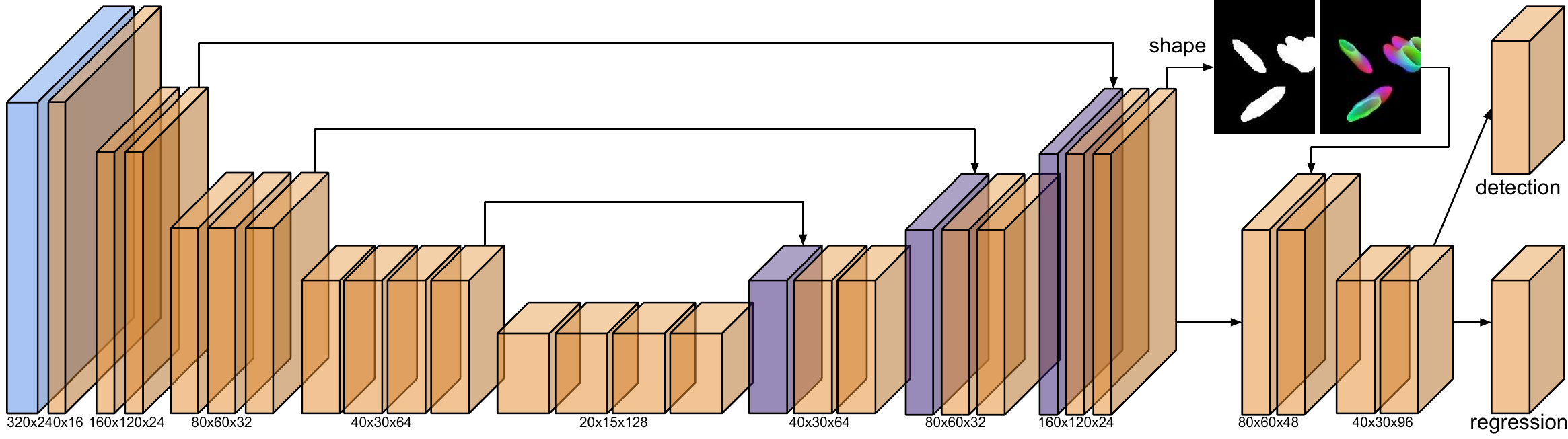}
    \caption{MobilePose-Shape network with shape prediction at an intermediate layer.}
    \label{fig:shape}
\end{figure}

Similar with previous methods~\cite{Hu_2019_Segmentation,Wang_2019_NOCS,Li_2019_CDPN,Park_2019_Pix2Pose}, we select coordinate map and segmentation as our intra-category shape features, as shown in Fig.~\ref{fig:shape}. The coordinate map has three channels, corresponding to the axes of 3D coordinates. If we have the CAD model of an object in training data, we can render coordinate map using normalize coordinates as colors. Coordinate map is a strong feature with pixel-level signals. However, it requires object's CAD model and pose, which are difficult to acquire. Therefore, we add segmentation as another shape feature. For simplicity, we use semantic segmentation, resulting in one additional channel in our shape supervision. Segmentation is a weak feature for pose estimation. That is, given a segmentation of an unseen object, it is not sufficient to determine its pose. Yet, it does not need object's CAD model and pose, and is easier to acquire.

\subsection{Shape-supervised Network}
With the shape features, we modify the network with high-resolution layers in the decoder and a shape prediction layer. In previous work~\cite{Hu_2019_Segmentation,Wang_2019_NOCS,Li_2019_CDPN,Park_2019_Pix2Pose}, they add another branch for shape prediction in parallel to other streams. The predicted shape is used out of network for building 2D-3D correspondence. In a contrary, we add an intermediate layer for shape prediction, whose output is further used in network. Meanwhile, shape prediction is useful in many applications, making our network a joint learning of multi-tasks: object detection, pose estimation, and shape prediction.

As shown in Fig.~\ref{fig:shape}, we combine multi-scale features in the decoder. A shape layer is added at the end of the decoder, predicting shape features. It is then concatenated with the decoder to connect pose heads after downsampling. Specifically, we utilize four inverted residual blocks to reduce the resolution, and finally attach the detection head and regression head, described in Section~\ref{sec:head}. The shape head ($160\times120\times4$) has four channels with L2 loss (mean squared error). Training examples without shape labels are skipped when computing this loss. Through experiments, we show that even with a weak supervision, the pose estimation is improved by introducing high-resolution shape prediction.

\section{Post-processing}\label{sec:post}
Despite the lightweight model, post-processing is another component that is critical to mobile applications. Expensive algorithms \eg RANSAC, large-sized PnP, and ICP, are not in our consideration. As a result, we simplify the post-processing to only two cheap operations: peak extraction and EPnP~\cite{Lepetit_2009_EPnP}.

To compute projected vertices of a 3D bounding box, we extract peaks of the detection output, a $40\times30$ heatmap. For a peak pixel $\mathbf{p}$, which may not necessarily be the center pixel, the eight vertices $\{\mathbf{x}_i\}$ of the projected bounding box can be simply computed by
\begin{equation}
 \mathbf{x}_i = \mathbf{p} + \mathcal{D}_i(\mathbf{p}), 
\end{equation}
where $\mathcal{D}_i(\mathbf{p})$ is the displacement field of vertex $\mathbf{x}_i$ computed by Eq.~\ref{eq:distance}.

Given the projected 2D box vertices and the camera intrinsics, we employ the EPnP~\cite{Lepetit_2009_EPnP} algorithm to recover a 3D bounding box up to scale. The algorithm has constant complexity, which solves eigen-decomposition of a $12\times12$ matrix. It does not require known object's size. We choose four control points $\{\mathbf{C}_j\}$ as the origin (at object's coordinate system), and three points along the coordinate axes. These control points form an orthogonal basis of the object frame. The eight vertices of a 3D bounding box can be represented by these four control points,
\begin{equation}\label{eq:control}
\mathbf{X}_i=\sum_{j=0}^{3}\alpha_{ij}\mathbf{C}_j,
\end{equation}
where $\{\alpha_{ij}\}$ are coefficients that are held under rigid transformations.

From camera projection, we obtain a linear system of 16 equations, with each box vertex contributing two equations for $u_i$ and $v_i$. By re-writing control points in camera frame as a 12-vector $\mathbf{C}^c$, this linear system can be formulated as
\begin{equation}
\mathbf{M}\cdot\mathbf{C}^c=0,
\end{equation}
where $\mathbf{M}$ is a $16\times12$ matrix composed by 2D vertices $\mathbf{x}_i$, camera intrinsics, and coefficients $\alpha_{ij}$. For details, please refer to our supplementary material. The solution to this linear system is the null eigenvectors of matrix $\mathbf{M}^T\mathbf{M}$. With this solution, we can recover a 3D bonding box in camera frame by Eq.~\ref{eq:control}, and further estimate object's pose and size up to a scale. 

\section{Training Data}
Since there is no existing large dataset of shoes with annotated poses, we build our dataset using on-device AR techniques. We annotate real data with 3D bounding boxes, and rely on synthetic data to provide weak supervision for shape.

\subsection{Real Data}
The lack of training data is a remaining challenge in 6DoF pose estimation. The majority of previous methods are instance-aware with full supervision from a small dataset. To solve the problem for unseen objects, we develop a pipeline to collect and annotate video clips recorded by mobile devices equipped with AR. The cutting-edge AR solutions (\eg ARKit and ARCore) can estimate camera poses and sparse 3D features on the fly using Visual Inertial Odometry (VIO). This on-device technology enables an affordable and scalable way to generate 3D training data.

The key to our data pipeline is efficient and accurate 3D bounding box annotation. We build a tool to visualize both 2D and 3D views of recorded sequences. Annotators draw 3D bounding boxes in the 3D view, and verify them in multiple 2D views across the sequence. The drawn bounding boxes are automatically populated to all frames in the video sequence, using estimated camera poses from AR. 

As a result, we annotated 1800 video clips of shoes. Clips are several-seconds long of different shoes under various environments. We only accepted one clip for one or a pair of shoes, and hence, the objects are completely different from clip to clip. Among the clips, 1500 were randomly selected for training, and the rest 300 were reserved for evaluation. Finally, considering adjacent frames from the same clip are very similar, we randomly selected 100K images for training, and 1K images for evaluation. As shown in Table~\ref{tab:data}, our real data only has 3D bounding box labels. This is because annotating pixel-level shape labels frame by frame is much more expensive.

\begin{table}[t]
\begin{center}
\begin{tabular}{l|c|c|c|c|c}\hline
Dataset & Pose & Segmentation & Coordinate Map & Size & \# of Objects \\
\hline
Real & annotation & N/A & N/A & 100K & 1500 \\
\hline
Synthetic-3D & accurate & accurate & accurate & 80K & 50 \\
\hline
Synthetic-2D & N/A & noisy & N/A & 50K & 75K\\
\hline
\end{tabular}
\caption{Datasets with their labels and sizes.}
\label{tab:data}
\end{center}
\end{table}

\subsection{Synthetic Data}

\begin{figure}[t]
    \centering
    \includegraphics[width=0.16\textwidth]{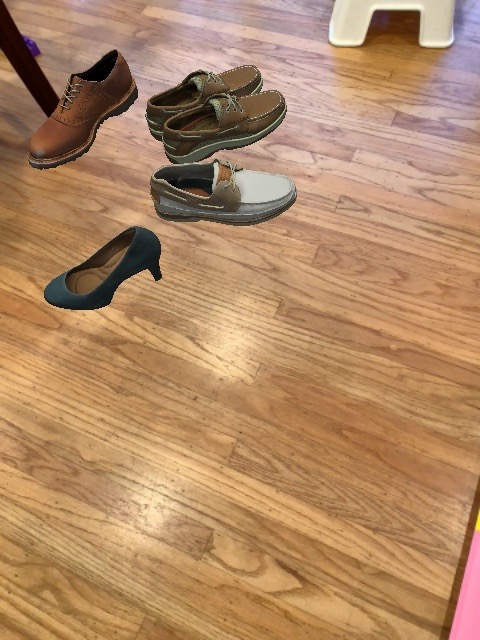}
    \includegraphics[width=0.16\textwidth]{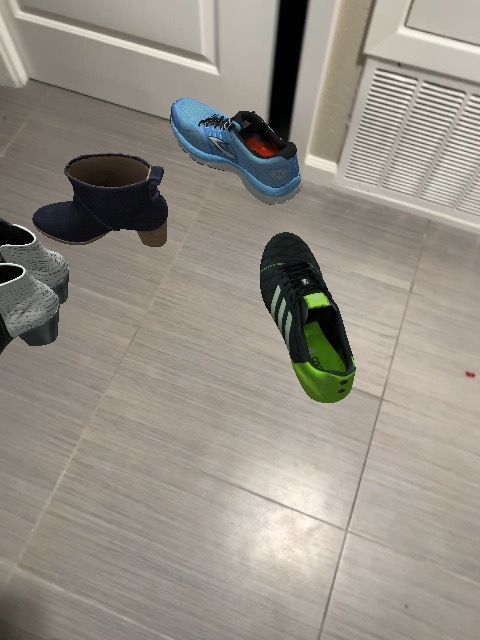}
    \includegraphics[width=0.16\textwidth]{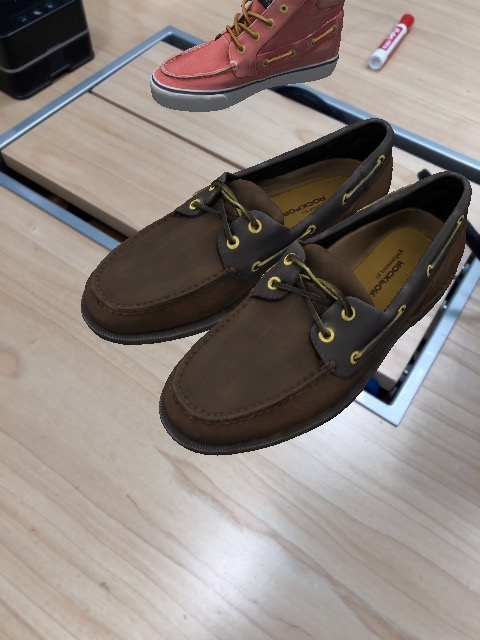}
    \includegraphics[width=0.16\textwidth]{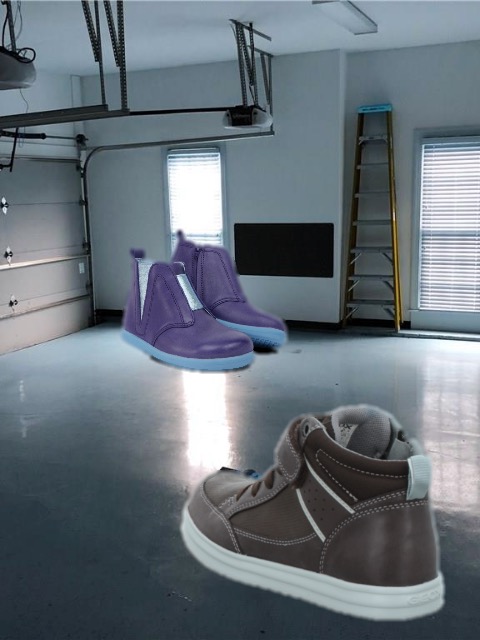}
    \includegraphics[width=0.16\textwidth]{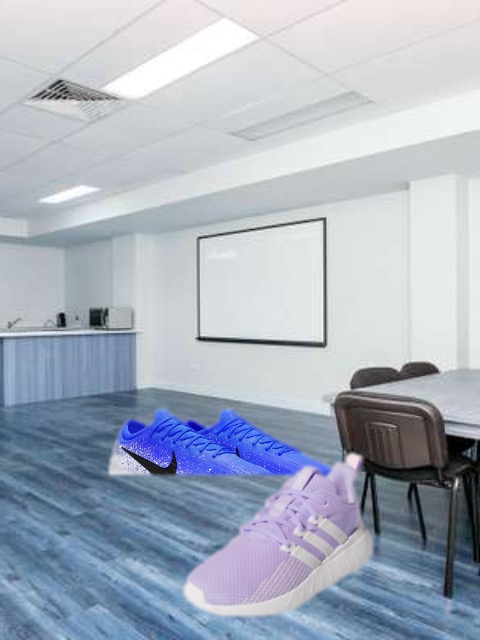}
    \includegraphics[width=0.16\textwidth]{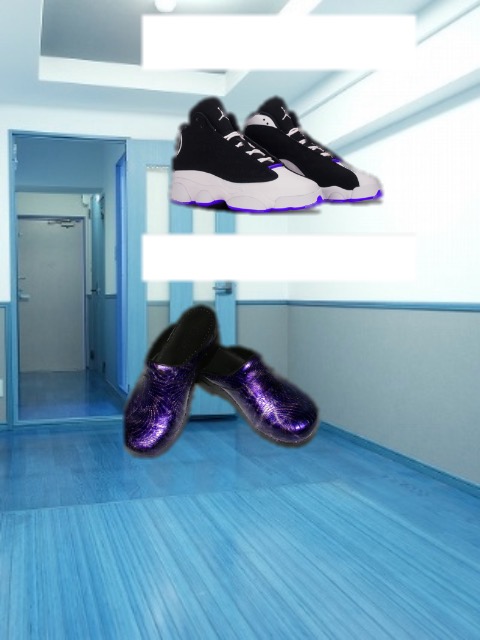}
    
    \includegraphics[width=0.16\textwidth]{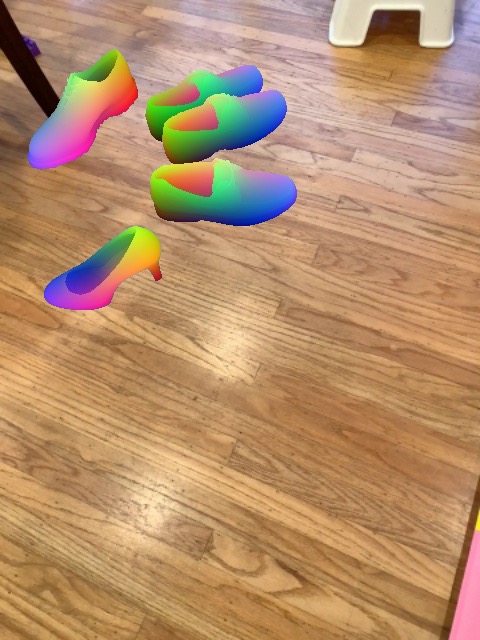}
    \includegraphics[width=0.16\textwidth]{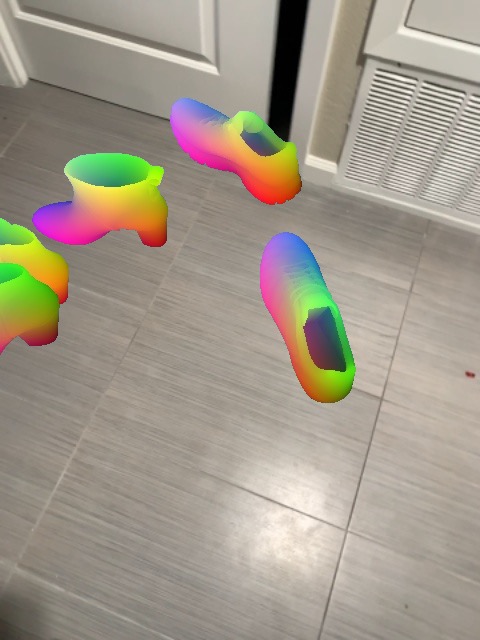}
    \includegraphics[width=0.16\textwidth]{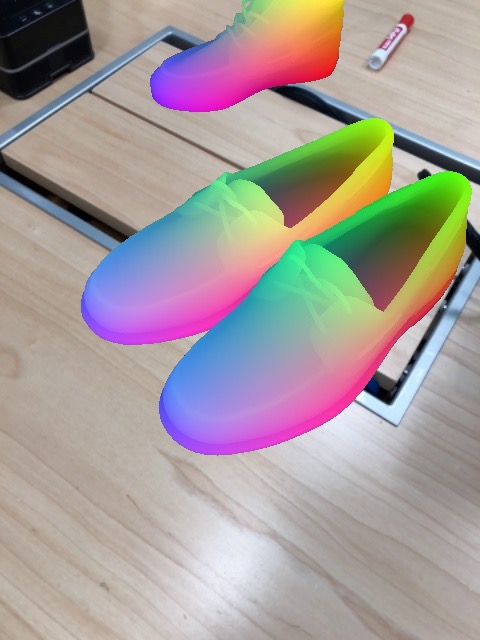}
    \includegraphics[width=0.16\textwidth]{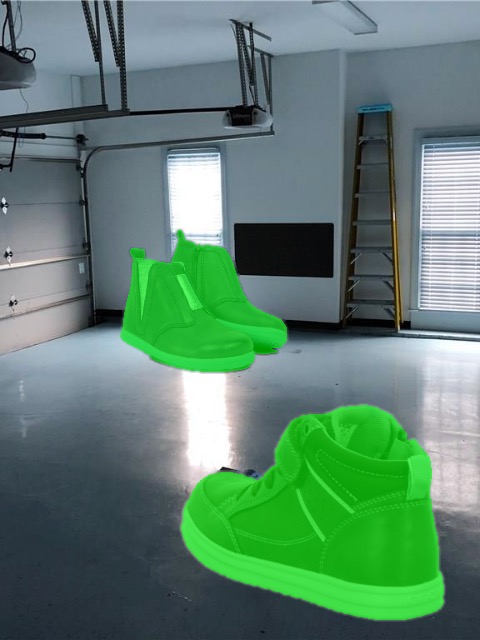}
    \includegraphics[width=0.16\textwidth]{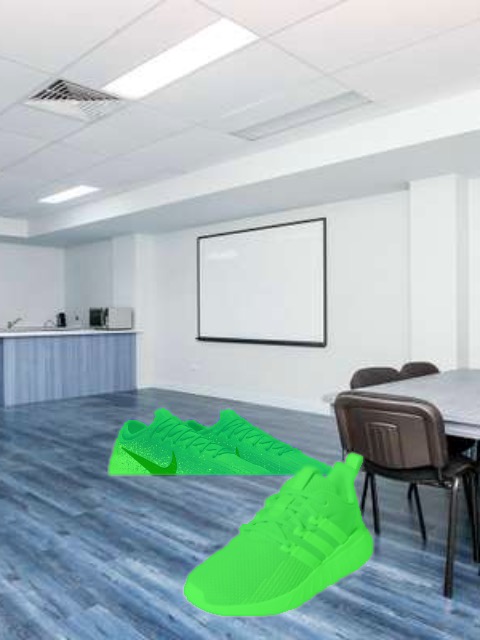}
    \includegraphics[width=0.16\textwidth]{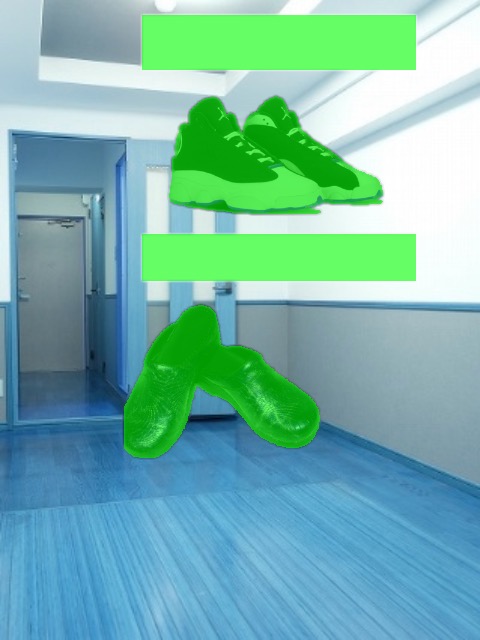}
    \caption{Examples of our synthetic-3D (left three columns), synthetic-2D (right three columns), and their shape labels (bottom row). The last two examples are considered with mild and severe label errors.}
    \label{fig:syn}
\end{figure}

To provide shape supervision and enrich the real dataset, we generate two sets of synthetic data. The first one synthetic-3D has 3D labels. We collect AR video clips of background scenes, and place virtual objects into the scenes. Specifically, we render virtual objects with random poses on a detected plane in the scene, \eg a table or a floor. We reuse estimated lights in AR sessions for lighting. Measurements in AR session data are in metric scale. Therefore, virtual objects are rendered coherently with the surrounding geometries. We collected 100 video clips of common scenes: home, office, and outdoor. For each scene, we generated 100 sequences by rendering 50 scanned shoes with random poses. Each sequence contains a number of shoes. From the generated images, we randomly selected 80K for training. As some examples shown in Fig.~\ref{fig:syn} the synthetic-3D data has 3D bounding box, segmentation, and coordinate map labels.

Although the synthetic-3D data has accurate labels, the numbers of objects and backgrounds are still limited. Therefore, we build a synthetic-2D dataset by images crawled from the internet. We crawled 75K images of shoes with transparent background, and 40K images of backgrounds (\eg office and home). Shoe images with trivial errors (\eg no transparent background) are filtered. We segmented shoes by the alpha channel, and randomly paste them to the background images. As shown in Fig.~\ref{fig:syn}, the generated images are not realistic, and the labels are noisy. We roughly estimated that there are about 20\% images with mild label errors (\eg small missing parts and shadows), and about 10\% images with severe label errors (\eg non-shoe objects and large extra background).

We summarize our datasets in Table~\ref{tab:data}. The three datasets are complementary to each other. The real data have real images collected at different places, the synthetic-3D data have accurate and complete labels, while the synthetic-2D data cover a large number of objects. With a low cost, we demonstrate a method to prepare training data with 2D and 3D labels, which can be used in other computer vision tasks.

\section{Experiments}

\subsection{Implementation Details and Evaluation Metric}
Our training pipeline is implemented in TensorFlow. We train our networks using the Adam optimizer with batch size 128 on 8 GPUs. The initial learning rate is 0.01 and gradually decays to 0.001 in 30 epochs. To deploy the trained models on mobile devices, we convert them to TFLite models. The conversion process will remove some layers such as batch normalization as they are not necessary during inference. Based on the MediaPipe framework~\cite{MediaPipe}, we build an application that runs on various mobile devices.

For evaluation metric, we adopt the average precision (AP) of 3D Intersection-over-Union (IoU). In previous work, the computation of 3D IoU is overly simplified. For example, \cite{Wang_2019_NOCS} assumes two 3D boxes are axis-aligned, and rotates one of them to get the best IoU. ~\cite{Tekin_2018_SingleShot} thought computing convex hull is tedious, and they used 2D IoU of projections. These simplifications do not hold in general 3D oriented boxes and often result in over-estimation of the 3D IoU metric. On the contrary, we compute the exact 3D IoU by finding intersecting points of two oriented boxes, and computing the volume of the intersection as a convex hull. Recall that our post-processing recovers 3D bounding boxes up to a scale. Although the scale is not necessary in our applications, it is needed in evaluation. We reuse the detected planes in AR session data to determine the metric scale of our estimations. Please refer to the supplementary material for details.

\begin{table}[t]
\begin{center}
\begin{tabular}{l@{\hskip 0.2cm}|@{\hskip 0.2cm}c@{\hskip 0.2cm}|@{\hskip 0.2cm}c@{\hskip 0.2cm}|@{\hskip 0.2cm}c@{\hskip 0.2cm}|c}
\hline
Scale & $20\times15$ & $40\times30$ & $80\times60$ & $160\times120$ \\
\hline
AP@0.5IoU & 0.5026 & \textbf{0.5343} & 0.5174 & 0.4916\\
\hline
\end{tabular}
\caption{Study of decoder scales on AP@0.5IoU.}
\label{tab:scale}
\end{center}
\end{table}

\begin{table}[t]
\begin{center}
\begin{tabular}{l|@{\hskip 0.3cm}c@{\hskip 0.3cm}|c|c|c}
\hline
Dataset & Base & Shape-No & Shape-CM & Shape-Full\\
\hline
Real & \textbf{0.5343} & 0.5102 & & \\
\hline
Real + Syn3D & 0.5541 & 0.5378 & \textbf{0.5793} & 0.5554 \\
\hline
Real + Syn3D + Syn2D & & & 0.5434 & \textbf{0.5939} \\
\hline
\end{tabular}
\caption{Study of datasets and network configurations on AP@0.5IoU.}
\label{tab:config}
\end{center}
\end{table}

\subsection{Ablation Studies}
We conduct ablation studies on decoder scale, shape supervision, and dataset. Table~\ref{tab:scale} shows the study on decoder scale of the MobilePose-Base. To compare different scales, we build the decoder as shown in Fig.~\ref{fig:shape}, and output at different levels. The detection and regression heads are also in the same experimenting scale. The result of this study shows that the MobilePose-Base achieves the best accuracy at scale $40\times30$, in terms of AP at 0.5 3D IoU. As the model has more layers with high-resolution features, its accuracy drops. This motivates us to look for shape supervision with pixel-level signals.

To study the contributions of our datasets and network components, we compare different configurations and document the results in Table~\ref{tab:config}. In this experiment, we compare four network configurations: MobilePose-Base (Base), MobilePose-Shape without shape supervision (Shape-No), MobilePose-Shape with coordinate map (CM) in shape supervision (Shape-CM), and the MobilePose-Shape with full shape supervision (Shape-Full). The configurations are evaluated by training on three data combinations by AP at 0.5 3D IoU.

On the real dataset, we observe that the MoblePose-Shape is no better than the MobilePose-Base. This is consistent with our experiment on decoder scales, that high-resolution features without supervision are not helpful in pose estimation. Trained on the real dataset and synthetic-3D dataset, the MobilePose-Shape with coordinate map as supervision has better performance than other configurations. This proves that shape supervision can improve pose estimation. We also notice that with full (segmentation + coordinate map) supervision, the accuracy goes down. This indicates that segmentation is a weak feature, which is redundant when coupling with coordinate map.

Finally, trained on the all three datasets, the MobilePose-Shape has the best performance with full supervision. Although synthetic-2D dataset only has noisy segmentation label, the network is able to evolve with this weak and noisy supervision. Segmentation makes a positive contribution when there are data with even noisy supervision.

\begin{figure}[t]
    \centering
    \includegraphics[width=0.16\textwidth]{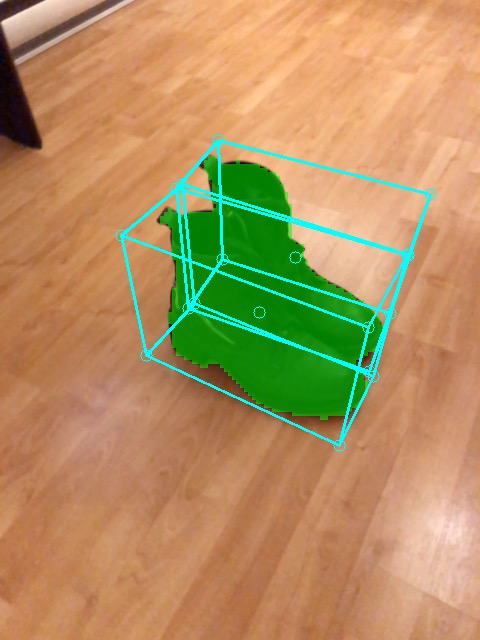}
    \includegraphics[width=0.16\textwidth]{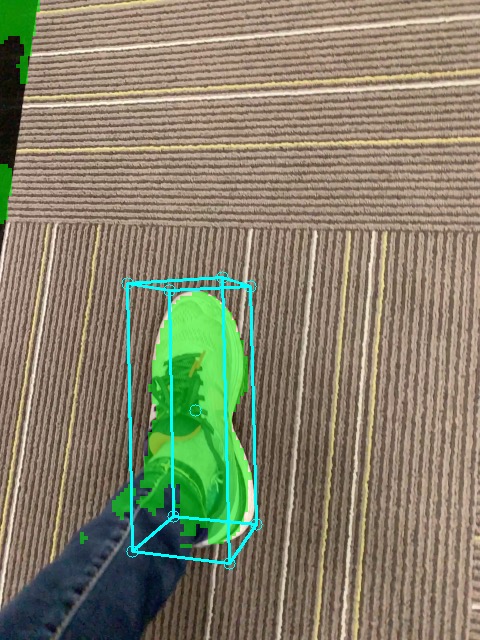}
    \includegraphics[width=0.16\textwidth]{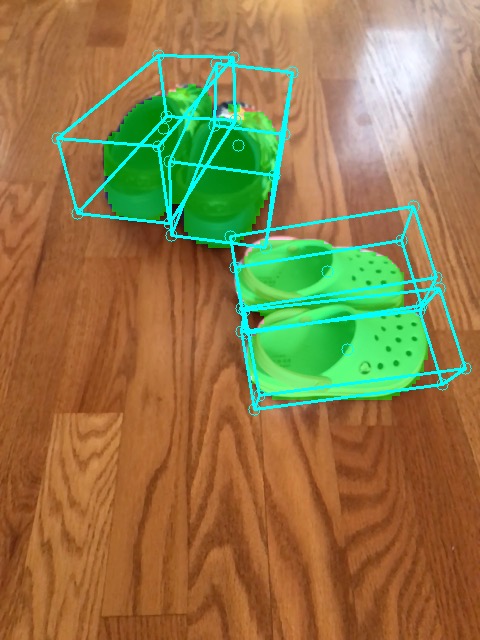}
    \includegraphics[width=0.16\textwidth]{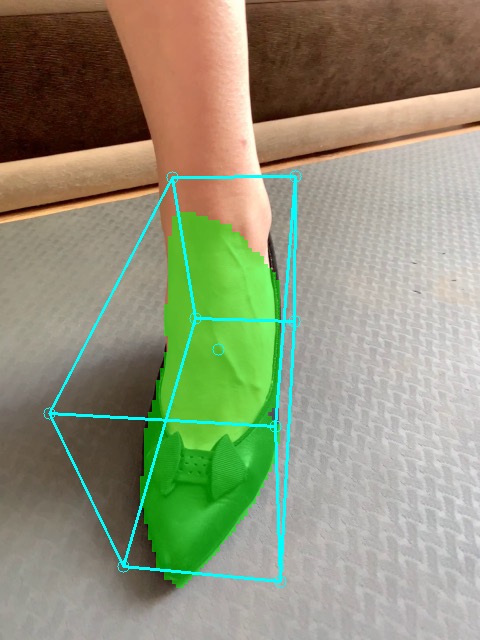}
    \includegraphics[width=0.16\textwidth]{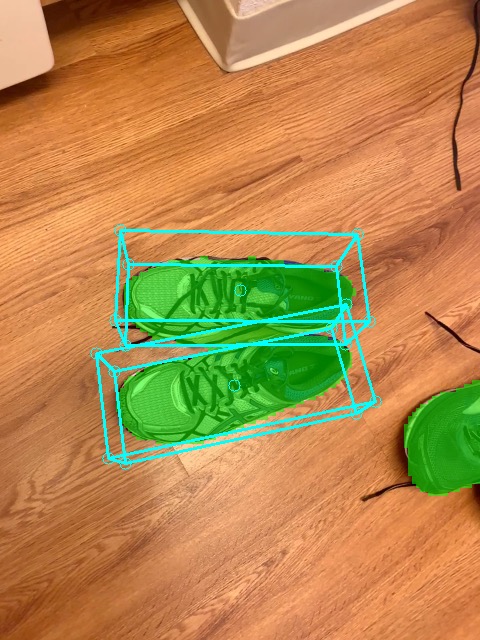}
    \includegraphics[width=0.16\textwidth]{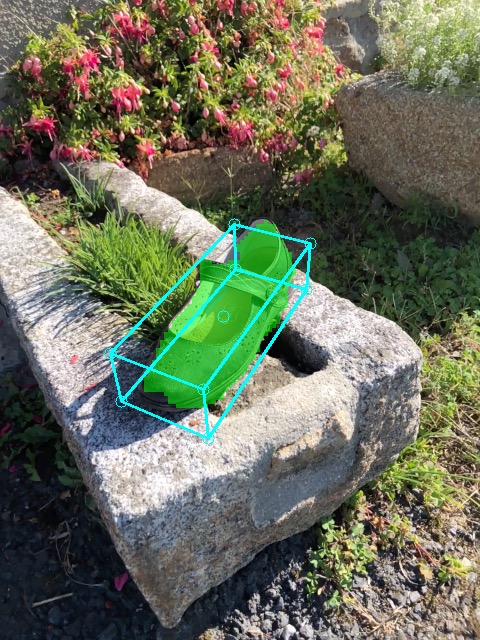}
    
    \includegraphics[width=0.16\textwidth]{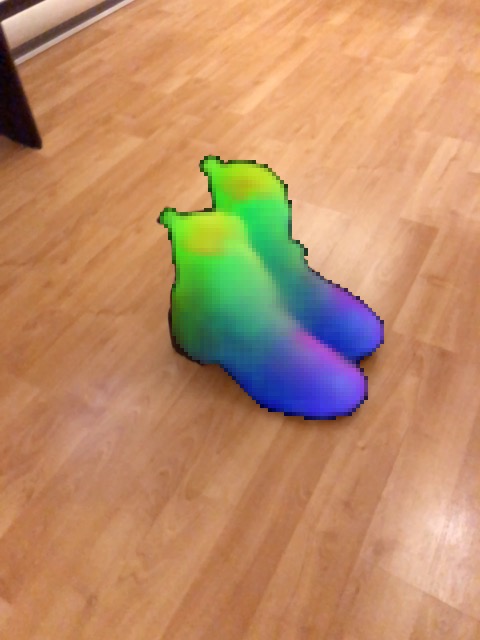}
    \includegraphics[width=0.16\textwidth]{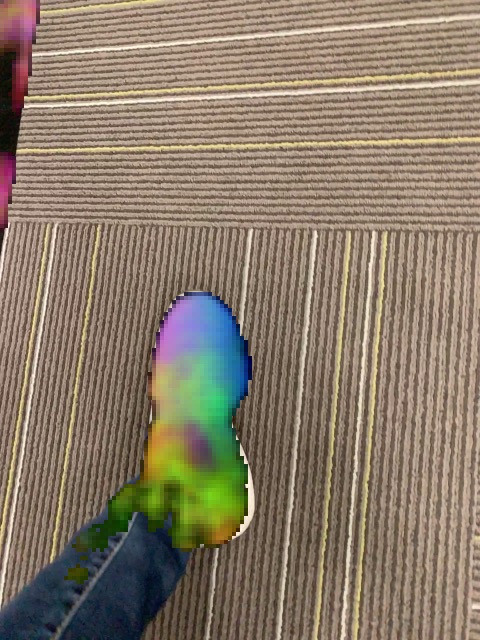}
    \includegraphics[width=0.16\textwidth]{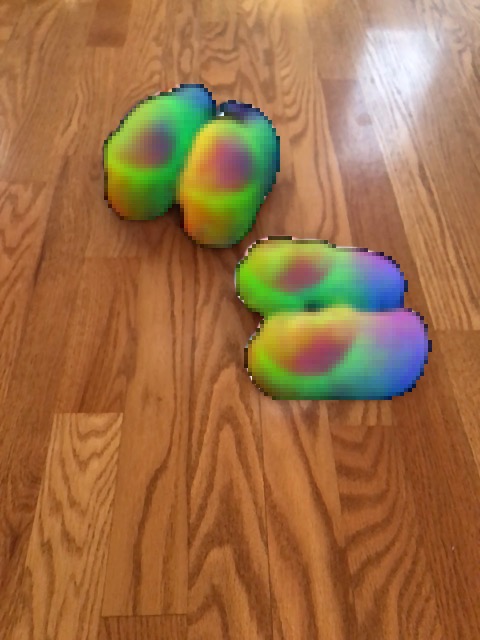}
    \includegraphics[width=0.16\textwidth]{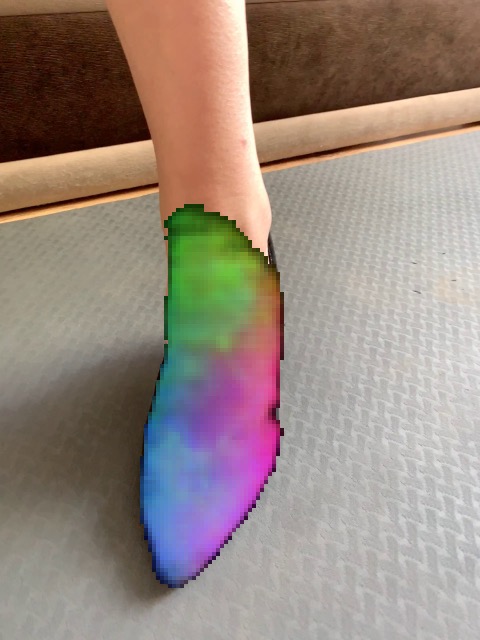}
    \includegraphics[width=0.16\textwidth]{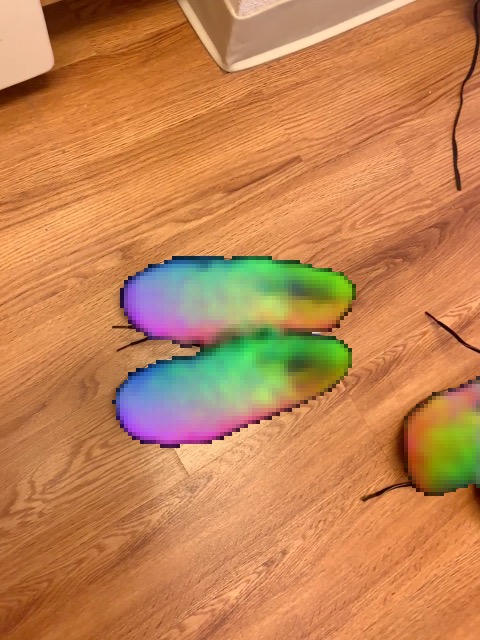}
    \includegraphics[width=0.16\textwidth]{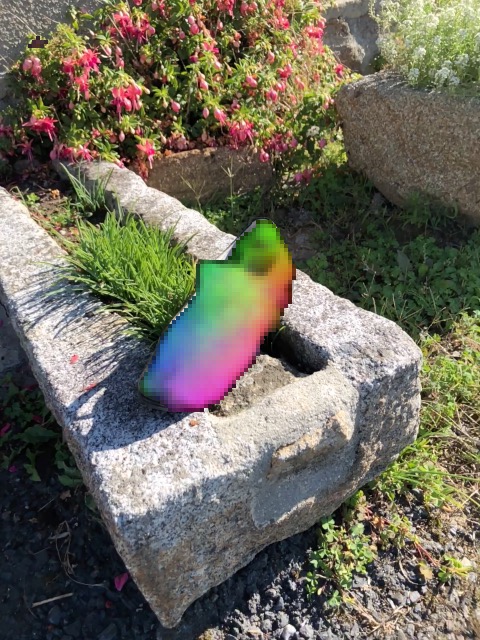}
    
    \includegraphics[width=0.16\textwidth]{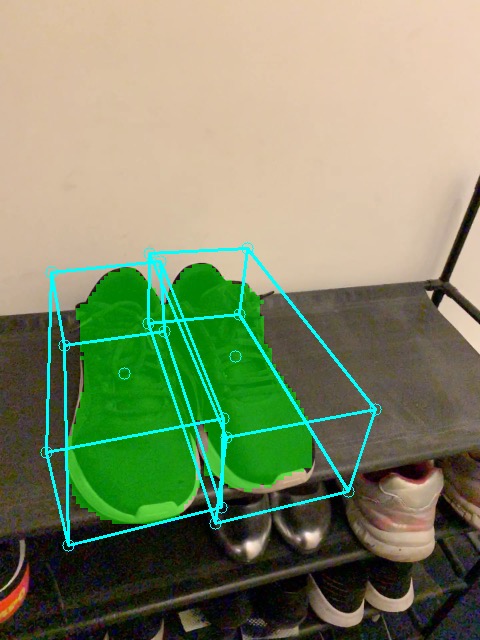}
    \includegraphics[width=0.16\textwidth]{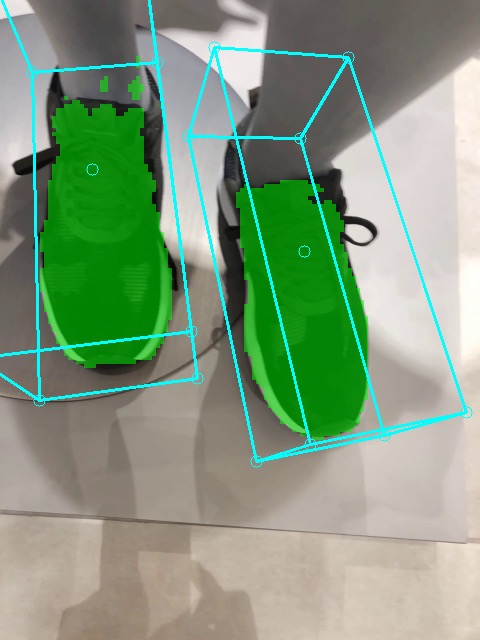}
    \includegraphics[width=0.16\textwidth]{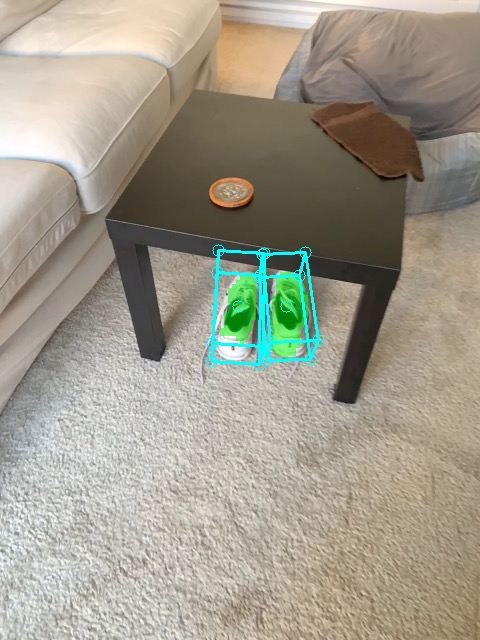}
    \includegraphics[width=0.16\textwidth]{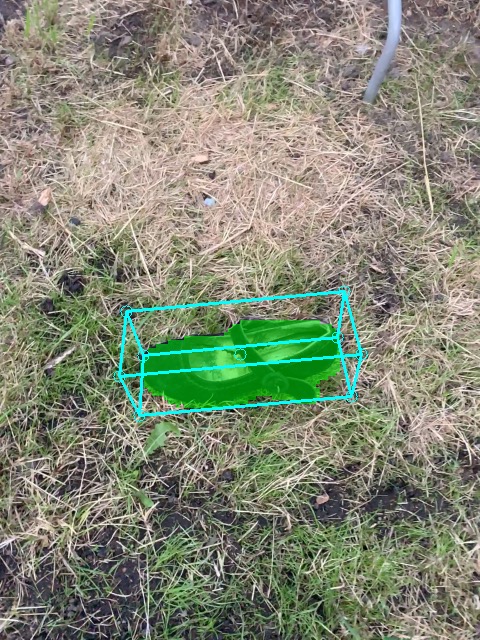}
    \includegraphics[width=0.16\textwidth]{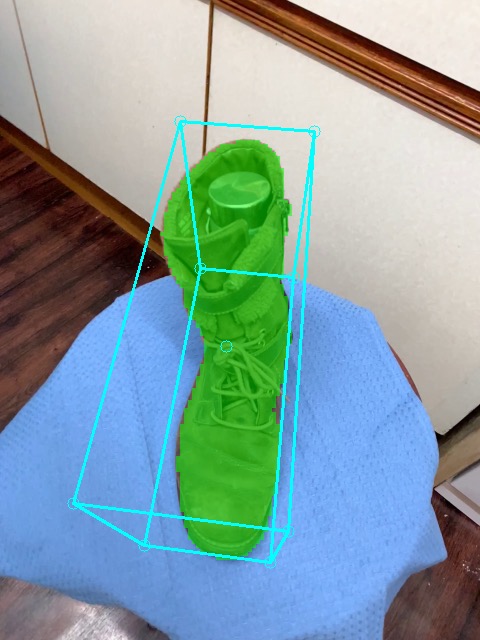}
    \includegraphics[width=0.16\textwidth]{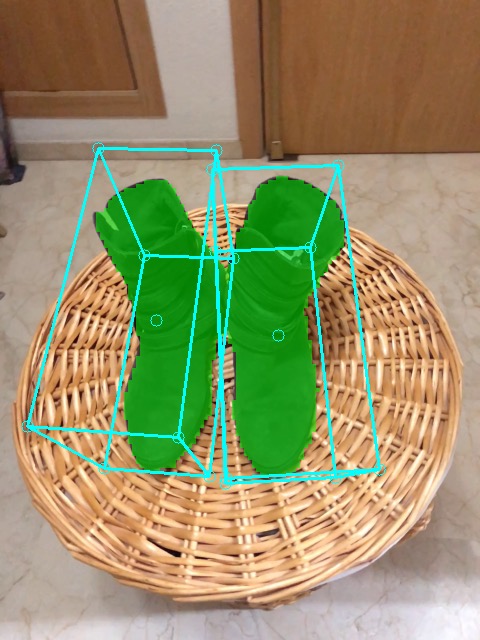}
    
    \includegraphics[width=0.16\textwidth]{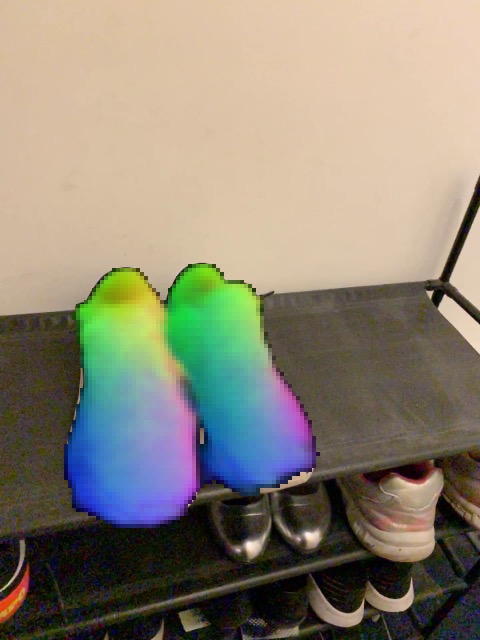}
    \includegraphics[width=0.16\textwidth]{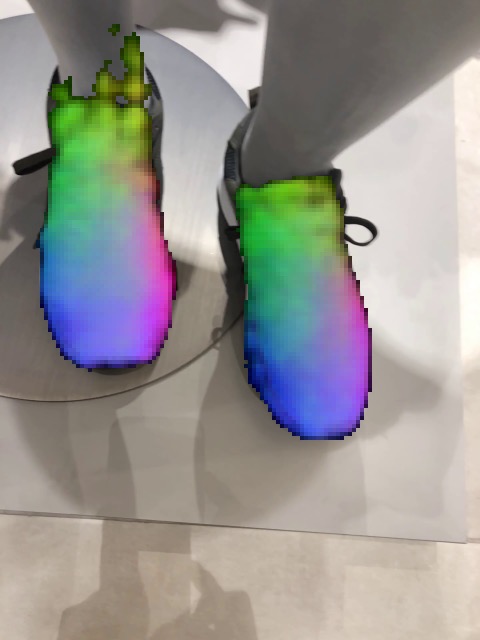}
    \includegraphics[width=0.16\textwidth]{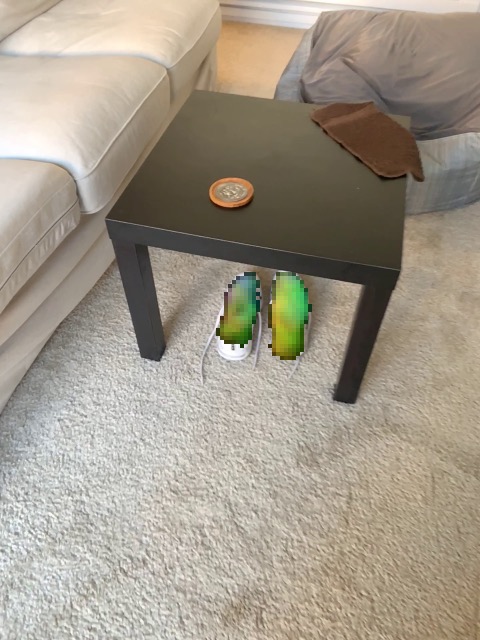}
    \includegraphics[width=0.16\textwidth]{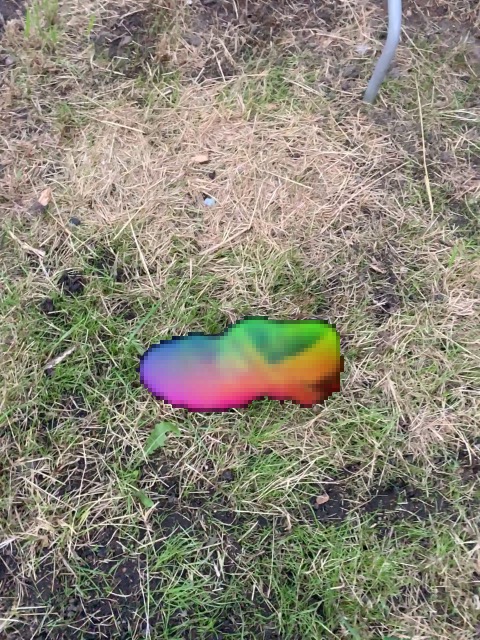}
    \includegraphics[width=0.16\textwidth]{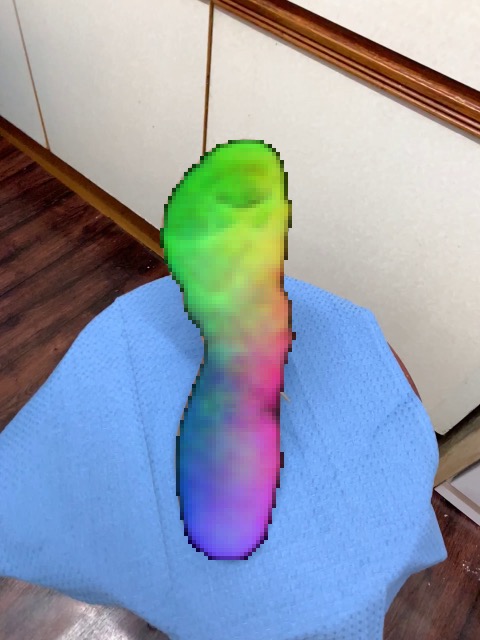}
    \includegraphics[width=0.16\textwidth]{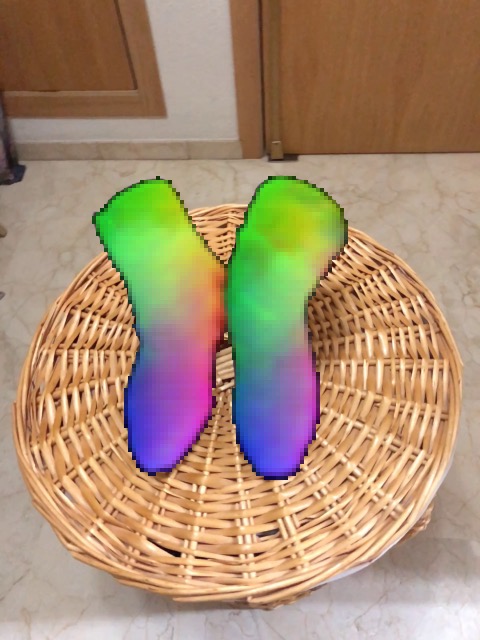}
    \caption{Example results on our real evaluation data. We show the reprojected bounding boxes of detected shoes, as well as the segmentation masks and coordinate maps learned from weak supervision.}
    \label{fig:result}
\end{figure}

\subsection{Results}

In Fig.~\ref{fig:result}, we show some results from our real evaluation dataset. To visualize the 3D bounding boxes, we reproject them to the 2D image plane. We also show the segmentation and coordinate map predictions from our model. We would like to remind readers that both segmentation and coordinate map are learned purely from synthetic data. Coordinate map only has 50 scanned shoes for supervision, while segmentation has very noisy labels. Recall that we argue learning accurate shape is more difficult than pose. We show that our model can infer accurate poses from weakly learned coarse shape features. Meanwhile, we show that our model also predicts reasonably well segmentation masks by transfer learning from synthetic data.

\begin{figure}[t]
    \centering
    \includegraphics[width=0.48\textwidth]{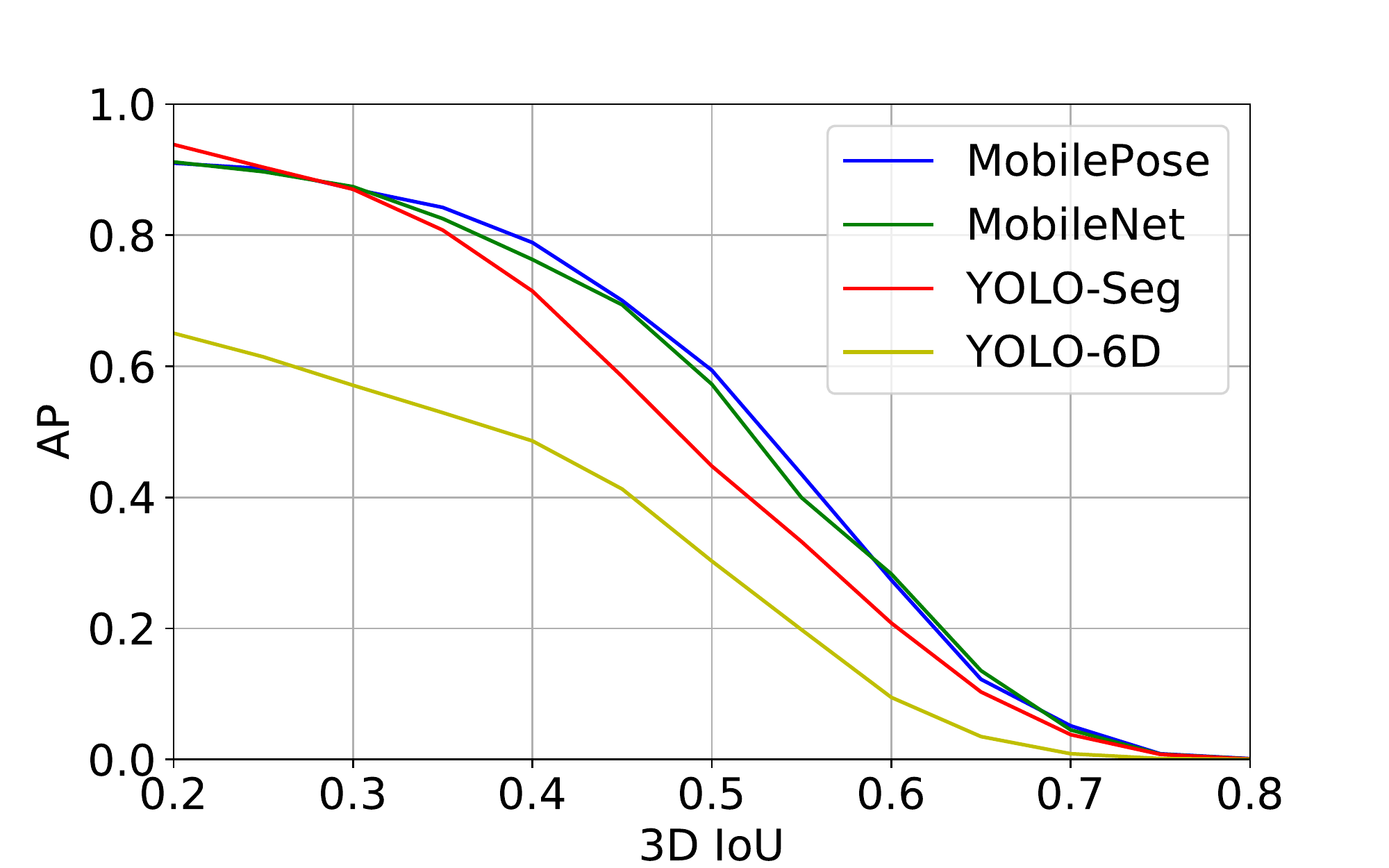}
    \includegraphics[width=0.36\textwidth]{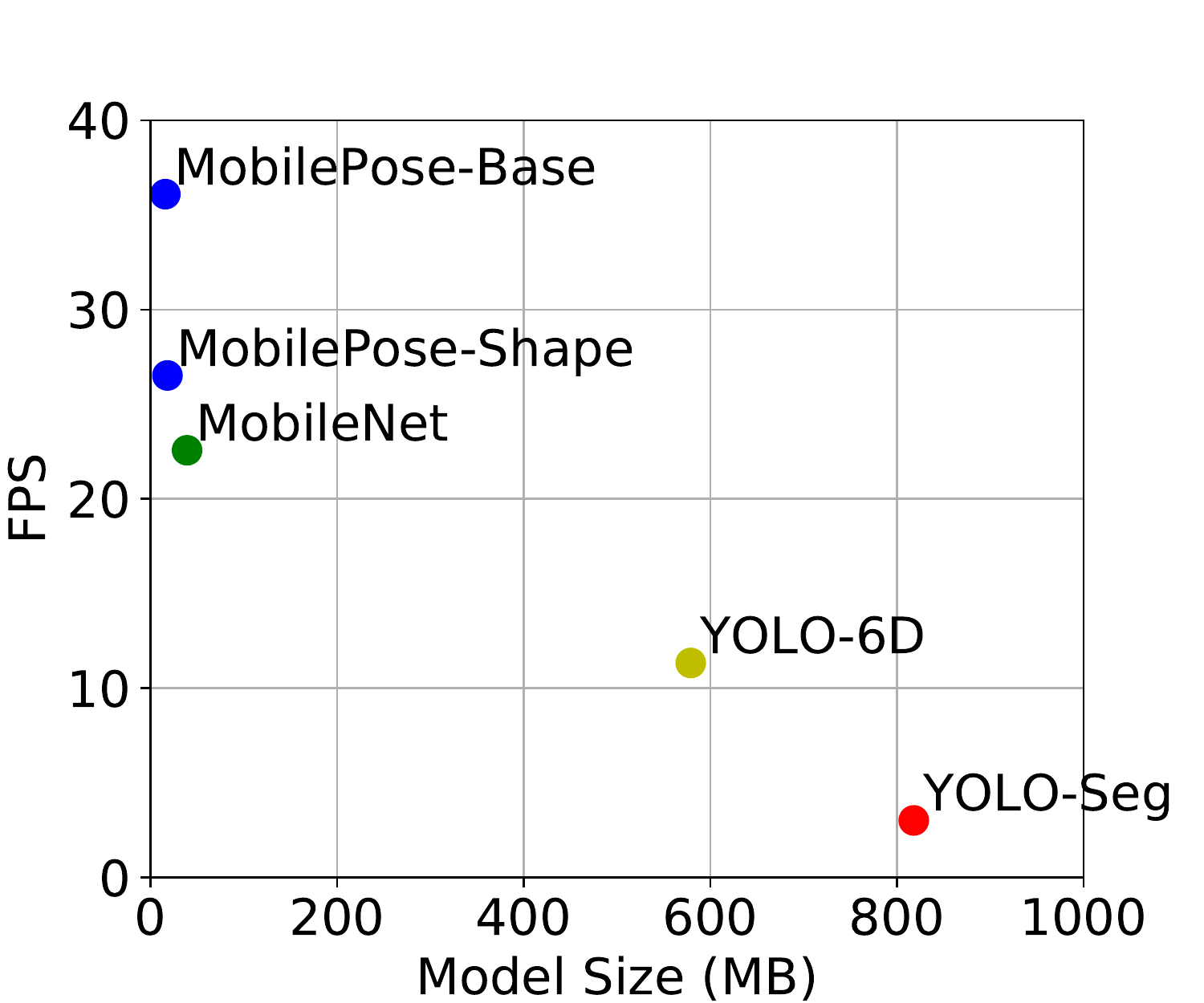}
    \caption{Comparison of state-of-the-art single-shot methods: AP vs 3D IoU (left), and FPS vs model size (right).}
    \label{fig:compare}
\end{figure}

\subsection{Comparisons}
\textbf{Shoe Dataset}. We compare our method with three other methods on our shoe dataset. Two of them are state-of-the-art single-shot solutions: YOLO-6D~\cite{Tekin_2018_SingleShot} and YOLO-Seg~\cite{Hu_2019_Segmentation}. Both of the two methods use YOLOv2~\cite{Redmon_2016_YOLOv2} as their backbones. The YOLO-6D predicts object's class and a confidence value at each grid cell, which are used for detecting the object. The YOLO-Seg uses a semantic segmentation branch for detection, parallel to its regression branch. Besides the two previous methods, we also compare with the MobileNetv2~\cite{Sandler_2018_MobileNetv2} attached with our decoder and heads. Recall that our encoder removes several blocks with large number of channels with nearly half parameters from MobileNetv2. This is to verify the effect of this optimization.


For all the methods, we follow the implementation details and uploaded code if there is any. As shown in Fig.~\ref{fig:compare}, our MobilePose-Shape has the best AP at 0.5 3D IoU. We also compare their model sizes and speeds on a smartphone (Galaxy S20 with Adreno 650 GPU). We benchmark models by running inference on mobile GPUs using TFLite with GPU delegate. Our MobilePose-Base runs at 36 FPS with the smallest model size (16MB), and MobilePose-Shape runs at 26.5 FPS with a slightly larger model size (18MB) and 10\% higher AP. Interestingly, using only half of the model size, our MobilePose-Shape is comparable with the MobileNetv2 plus our decoder and heads. This indicates that by introducing high-resolution features with even a weak supervision, low-resolutions features can be compressed by using a shallower and thinner network. Finally, comparing with the two previous models, ours are about $2\sim3\%$ in model size or number of parameters, and $3\sim12$ times in FPS.

\textbf{Public Datasets}. We also compare the methods on two popular public datasets: Linemod~\cite{Hinterstoisser_2012_LINEMOD} and Occlusion~\cite{Brachmann_2014_Occlusion}. We adopt the standard metrics of reprejection error (REP-5px) and average distance (ADD-0.1d), same as~\cite{Tekin_2018_SingleShot,Hu_2019_Segmentation}. The results are shown in Table~\ref{tab:linemod} and~\ref{tab:occlusion}, where superscript $*$ indicates the object is symmetric. We use our MobilePose-Base for the two experiments, because objects in the two datasets are relatively small. Without increasing model size with higher resolution, or leveraging detection-aware cropping, shape features do not provide sufficient supervision. Since our model is designed for unseen objects from a single category, we trained a model for each object category. Comparing with YOLO-Seg~\cite{Hu_2019_Segmentation} that uses a multi-object model for Occlusion, ours has much higher accuracy, and the total model size is still smaller.

\begin{table}[t]
\begin{center}
\begin{tabular}{c|c||c|c|c|c|c|c|c|c}\hline
Metric & Method & Ape & Can & Cat & Driller & Duck & Eggbox$^*$ & Glue$^*$ & Holep.\\
\hline
\multirow{2}{*}{REP-5px} &
YOLO-6D &  92.10 & \textbf{97.44} & 97.41 & 79.41 & 94.65 & 90.33 & \textbf{96.53} & 92.86 \\
&MobilePose & \textbf{98.92} & 95.56 & \textbf{99.44} & \textbf{85.47} & \textbf{97.86} & \textbf{99.47} & 95.63 & \textbf{97.85}\\
\hline
\multirow{2}{*}{ADD-0.1d} &
YOLO-6D & 21.62 & 68.80 & 41.82 & 63.51 & 27.23 & 69.58 & 80.02 & 42.63 \\
&MobilePose & \textbf{42.70} & \textbf{72.78} & \textbf{50.28} & \textbf{68.16} & \textbf{39.57} & \textbf{91.98} & \textbf{93.44} & \textbf{56.45} \\
\hline
\end{tabular}
\caption{Comparison with YOLO-6D~\cite{Tekin_2018_SingleShot} on Linemod dataset.}
\label{tab:linemod}
\end{center}
\end{table}

\begin{table}[t]
\begin{center}
\vspace{-0.5cm}
\begin{tabular}{c|c||c|c|c|c|c|c|c|c}\hline
Metric & Method & Ape & Can & Cat & Driller & Duck & Eggbox$^*$ & Glue$^*$ & Holep.\\
\hline
\multirow{2}{*}{REP-5px} &
YOLO-Seg & 59.1 & 59.8 & 46.9 & 59.0 & 42.6 & 11.9 & 16.5 & 63.6 \\
&MobilePose & \textbf{95.9} & \textbf{87.9} & \textbf{89.8} & \textbf{84.1} & \textbf{86.0} & \textbf{72.1} & \textbf{54.6} & \textbf{88.0}\\
\hline
\multirow{2}{*}{ADD-0.1d} &
YOLO-Seg & 12.1 & 39.9 & 8.2 & 45.2 & 17.2 & 22.1 & 35.8 & 36.0 \\
&MobilePose & \textbf{29.0} & \textbf{56.0} & \textbf{33.5} & \textbf{70.3} & \textbf{25.6} & \textbf{55.2} & \textbf{58.5} & \textbf{48.1} \\
\hline
\end{tabular}
\caption{Comparison with YOLO-Seg~\cite{Hu_2019_Segmentation} on Occlusion dataset.}
\label{tab:occlusion}
\end{center}
\end{table}

\section{Conclusion}
In this paper, we address the problem of pose estimation from two different angles that have not been explored before. First, we do not assume any prior knowledge of the unseen objects. Second, our MobilePose models are ultra lightweight that can run on mobile devices in real-time. Additionally, we reveal that pixel-level shape supervision can guide the network to learn poses from high-resolution features. With a particular interest, we demonstrate our models on various unseen shoes through a mobile application. Furthermore, the proposed method can be easily extended to other object categories.

%
%
\bibliographystyle{splncs04}

\end{document}